# Object Recognition from very few Training Examples for Enhancing Bicycle Maps

Christoph Reinders[1], Hanno Ackermann[1], Michael Ying Yang[2], and Bodo Rosenhahn[1]

*Abstract*— In recent years, data-driven methods have shown great success for extracting information about the infrastructure in urban areas. These algorithms are usually trained on large datasets consisting of thousands or millions of labeled training examples. While large datasets have been published regarding cars, for cyclists very few labeled data is available although appearance, point of view, and positioning of even relevant objects differ. Unfortunately, labeling data is costly and requires a huge amount of work.

In this paper, we thus address the problem of learning with very few labels. The aim is to recognize particular traffic signs in crowdsourced data to collect information which is of interest to cyclists. We propose a system for object recognition that is trained with only 15 examples per class on average. To achieve this, we combine the advantages of convolutional neural networks and random forests to learn a patch-wise classifier. In the next step, we map the random forest to a neural network and transform the classifier to a fully convolutional network. Thereby, the processing of full images is significantly accelerated and bounding boxes can be predicted. Finally, we integrate data of the Global Positioning System (GPS) to localize the predictions on the map. In comparison to Faster R-CNN and other networks for object recognition or algorithms for transfer learning, we considerably reduce the required amount of labeled data. We demonstrate good performance on the recognition of traffic signs for cyclists as well as their localization in maps.

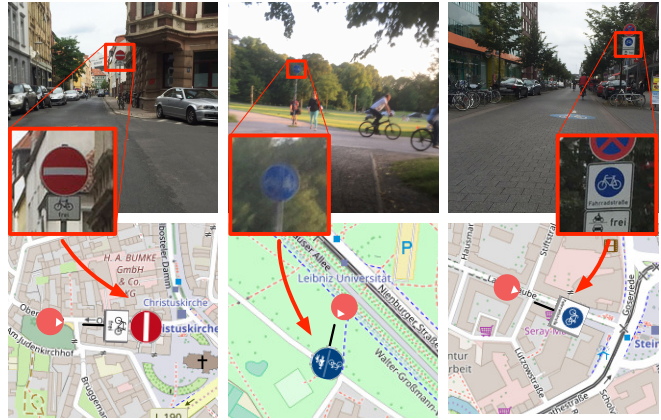

Fig. 1: Real-world data has great potential to provide traffic information that is of interest to cyclists. For example, roads that are prohibited for cars but free for cyclists (left), bicycle lines in parks (middle), or bicycle boulevards which are optimized for cyclists (right). All three examples are recognized by our system.

## I. INTRODUCTION

Cycling as a mode of transport has attracted growing interest. Cities are transforming urban transportation to improve their infrastructure. While current development shows more and more infrastructure improvements, road conditions can vary greatly. Cyclists are frequently confronted with challenges such as absence of bicycle lanes, being overlooked by cars, or bad roads. Arising safety concerns represent a barrier for using bicycles. Thus, recommending fast and safe routes for cyclists has great potential in terms of environmental and mobility aspects. This, in turn, requires detailed information about roads and traffic regulations.

For cars, precise information has become available. Google, for example, started the Google Street View project in which data is captured by many cars. These are equipped with stereo cameras which already offer a good 3D estimation in a certain range, lidar, and other sensors. Additionally the cars provide computational power as well as power supply. In research, popular datasets like GTSRB [1], KITTI [2], and Cityscapes [3] have been published.

In recent years, users are increasingly involved in the data collection. Crowdsourcing data enables to create large amount of real-world datasets. For example, the smart phone app Waze collects data such as GPS-position and speed from multiple users to predict traffic jams. OpenStreetMap aims to build a freely available map of the world to which users can easily contribute.

Machine learning techniques have shown great success for analyzing this data. While large amounts of data can be quickly collected, supervised learning further requires labeled data. Labeling data, unfortunately, is usually very time-consuming and literally expensive.

Our motivation is to collect information which is of interest to cyclists. Analyzing street data for cyclists cannot be straightforwardly done by using data captured for cars due to different perspectives, different street signs, and routes prohibited for cars but not for bicycles, as shown in Fig. 1. For collecting real-world data, we involve users by using smart phones that are attached to their bicycles. Compared to other systems like for example Google Street View our recording system consists of a single consumer camera and can only rely on a limited power supply as well as little computational power. On the other hand, our system has very low hardware costs and is highly scalable so that

[1]Institute for Information Processing, Leibniz University Hanover, 30167 Hanover, Germany {reinders, ackermann, rosenhahn}@tnt.uni-hannover.de

[2]Scene Understanding Group, University of Twente, Netherlands michael.yang@utwente.nl

This research was supported by German Research Foundation DFG within Priority Research Program 1894 Volunteered Geographic Information: Interpretation, Visualization and Social Computing.

crowdsourcing becomes possible.

Although capturing data becomes easy with this system, generating labels is still very expensive. Thus, in this paper we further address the problem of learning with extremely little data to recognize traffic signs relevant for cyclists. We combine multiple machine learning techniques to create a system for object recognition. Convolutional neural networks (CNN) have shown to learn strong feature representations. On the other hand, random forests (RF) achieve very good results in regression and classification tasks even when few data is available. To combine both advantages we generate a feature extractor using a CNN and train a random forest based on the features. We map the random forest to a neural network and transform the full pipeline into a fully convolutional network. Thus, due to the shared features, the processing of full images is significantly accelerated. The resulting probability map is used to perform object detection. In a next step, we integrate information of a GPS-sensor to localize the detections on the map.

To summarize, our **contributions** are:
- We propose a pipeline for training a traffic sign recognition system based on convolutional neural networks, using only **15 training samples per class** on average.
- We integrate GPS-information to localize the predicted traffic signs on the map.
- We collected a dataset of images of street scenes from the perspective of cyclists by crowdsourcing. The images are captured using a mobile device that is attached to the bicycle.
- Our recorded data is preprocessed directly on the mobile device. Meta data is used to keep interesting images to minimize redundancy and the amount of data.
- We publish the training and test dataset for traffic sign detection[1].

## II. RELATED WORK

In recent years, convolutional neural networks have become the dominant approach for various tasks including classification [4], object recognition, and scene analysis [5], [6]. Girshick *et al.* [7] proposed a multi-stage pipeline called Regions with Convolutional Neural Networks (R-CNN) for the classification of region proposals to detect objects. It achieves good results but the pipeline is less efficient because features of each region proposal need be computed repeatedly. In SPP-net [8], this problem has been addressed by introducing a pooling strategy to calculate the feature map only once and generate features in arbitrary regions. Fast R-CNN [9] further improves the speed and accuracy by combining multiple stages. A drawback of these algorithms is their large dependence on the region proposal method. Faster R-CNN [10] combines the region proposal mechanism and a CNN classifier within a single network by introducing a Region Proposal Network. Due to shared convolutions, region proposals are generated at nearly no extra cost. Other networks such as SSD [11] directly regress bounding boxes without generating object proposals in an end-to-end network. YOLO [12] is a similar approach which is extremely fast but comes with some compromise in detection accuracy. Generally, these networks perform very well. However, they typically consist of millions of variables and for estimating those, a large amount of labeled data is required for training.

Feature learning and transferring techniques have been applied to reduce the required amount of labeled data [13]. The problem of insufficient training data has also been addressed by other works such as [14] and [15]. Moysset *et al.* [14] proposed a new model that predicts the bounding boxes directly. Wagner *et al.* [15] compared unsupervised feature learning methods and demonstrated performance boosts by pre-training. Although transfer learning techniques are applied, the networks still have a large number of variables for fine-tuning.

A different approach is the combination of random forests and neural networks. Deep Neural Decision Forests [16] unifies both in a single system that is trained end-to-end. Sethi [17] and Welbl [18] presented a mapping of random forests to neural networks. The mapping can be used for several applications. Massiceti *et al.* [19] demonstrated the application for camera localization. Richmond *et al.* [20] explored the mapping of stacked RFs to CNNs and an approximate mapping back to perform semantic segmentation.

## III. TRAFFIC SIGN RECOGNITION

In this section, we present a system for recognizing traffic signs. To overcome the problem of lack of data, we first build a classifier that predicts the class probabilities of a single image patch. This is done in two steps. First, we train a CNN on a different dataset where large amount of data is available. Afterwards we use the generated features, extract the feature vectors, and train a random forest. The resulting classifier can be used to perform patch-wise prediction and to build a probability map for a given full image. Subsequently, all traffic signs are extracted and the recognition system outputs the class as well as the corresponding bounding box.

Finally, the processing of full images is accelerated. By mapping the random forest to a neural network, it becomes possible to combine feature generation and classification. Afterwards we transform the neural network to a fully convolutional network.

### A. Feature Learning

We learn features by training a convolutional neural network $\text{CNN}^\text{F}$. The patch size is $32 \times 32$. We adopt the network architecture of Springenberg *et. al* [21]. To reduce the memory requirements, we decrease the number of filters in *conv1* to 32, in *conv2* to 64, and in *conv3* to 128.

Because we have only few labeled data available, we train the network on the larger dataset GTSRB [1]. After training, the resulting network $\text{CNN}^\text{F}$ can be used to generate feature vectors by passing an input image to the network and performing a forward pass. The feature vectors can be extracted from the last convolutional layer. In our network

---
[1]www.tnt.uni-hannover.de/~reinders/

this corresponds to the third convolutional layer, denoted by $\text{CNN}^{\text{F}}_{relu3}(x)$.

*B. Random Forest Classification*

Usually, neural networks perform very good in classification. However, if the data is limited, the large amount of parameters to be trained causes overfitting. Random forests [22] have shown to be robust classifiers even if few data is available. A random forest consists of multiple decision trees. Each decision tree uses a randomly selected subset of features and training data. The output is calculated by averaging the individual decision tree predictions.

After creating a feature generator, we calculate the feature vector $f^{(i)} = \text{CNN}^{\text{F}}_{relu3}(x^{(i)})$ for every input vector $x^{(i)}$. Based on the feature vectors we train a random forest that predicts the target values $y^{(i)}$. By combining the feature generator $\text{CNN}^{\text{F}}$ and the random forest, we construct a classifier that predicts the class probabilities for an image patch. This classifier can be used to process a full input image patch-wisely. Calculating the class probabilities for each image patch produces an output probability map.

*C. RF to NN Mapping*

Here, we present a method for mapping random forests to two-hidden-layer neural networks introduced by Sethi [17] and Welbl [18]. The mapping is illustrated in Fig. 2.

A decision tree consists of *split nodes* $\mathcal{N}^{\text{Split}}$ and *leaf nodes* $\mathcal{N}^{\text{Leaf}}$. Each split node $s \in \mathcal{N}^{\text{Split}}$ performs a split decision and routes a data sample $x$ to the left child node $\text{cl}(s)$ or to the right child node $\text{cr}(s)$. When using axis-aligned split decisions the split rule is based on a single split feature $f(s)$ and a threshold value $\theta(s)$:

$$x \in \text{cl}(s) \iff x_{f(s)} < \theta(s) \quad (1)$$
$$x \in \text{cr}(s) \iff x_{f(s)} \geq \theta(s). \quad (2)$$

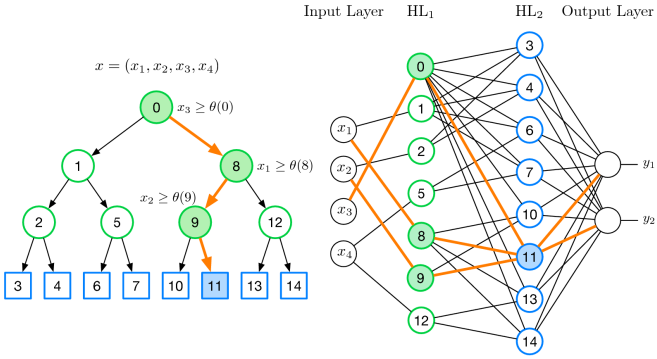

Fig. 2: A decision tree (left) and the mapped neural network (right). Each *split node* in the tree – indicated as circle – creates a neuron in the first hidden layer which evaluates the split rule. Each *leaf node* – indicated as rectangle – creates a neuron in the second hidden layer which determines the leaf membership. For example, a routing to leaf node 11 involves the split nodes (0, 8, 9). The relevant connections for the corresponding calculation in the neural network are highlighted.

All leaf nodes $l \in \mathcal{N}^{\text{Leaf}}$ store votes for the classes $y^l = (y_1^l, \ldots, y_C^l)$, where $C$ is the number of classes. For each leaf a unique path $\text{P}(l) = (s_0, \ldots, s_d)$ from root node $s_0$ to leaf $l$ over a sequence of split nodes $\{s_i\}_{i=0}^d$ exists, with $l \subseteq s_d \subseteq \cdots \subseteq s_0$. By evaluating the split rules for each split node along the path $\text{P}(l)$ the leaf membership can be expressed as:

$$x \in l \iff \forall s \in \text{P}(l) : \begin{cases} x_{f(s)} < \theta(s) & \text{if } l \in \text{cl}(s) \\ x_{f(s)} \geq \theta(s) & \text{if } l \in \text{cr}(s) \end{cases} \quad (3)$$

**First Hidden Layer.** The first hidden layer computes all split decisions. It is constructed by creating one neuron $\text{H}_1(s)$ per split node evaluating the split decision $x_{f(s)} \geq \theta(s)$. The activation output of the neuron should approximate the following function:

$$\text{a}(\text{H}_1(s)) = \begin{cases} -1, & \text{if } x_{f(s)} < \theta(s) \\ +1, & \text{if } x_{f(s)} \geq \theta(s) \end{cases} \quad (4)$$

where $-1$ encodes a routing to the left child node and $+1$ a routing to the right child node. Therefor the $f(s)^{\text{th}}$ neuron of the input layer is connected to $\text{H}_1(s)$ with weight $w_{f(s),\text{H}_1(s)} = \text{c}_{01}$, where $\text{c}_{01}$ is a constant. The bias of $\text{H}_1(s)$ is set to $b_{\text{H}_1(s)} = -\text{c}_{01} \cdot \theta(s)$. All other weights are zero. As a result, the neuron $\text{H}_1(s)$ calculates the weighted input

$$\text{c}_{01} \cdot x_{f(s)} - \text{c}_{01} \cdot \theta(s) \quad (5)$$

which is smaller than zero when $x_{f(s)} < \theta(s)$ is fulfilled and greater or equal to zero otherwise. We use $\tanh(\cdot)$ as activation.

**Second Hidden Layer.** The second hidden layer combines the split decisions from layer $\text{H}_1$ to indicate the leaf membership $x \in l$. One leaf neuron $\text{H}_2(l)$ is created per leaf node. It is connected to all split neurons $\text{H}_1(s)$ along the path $s \in \text{P}(l)$ as follows

$$w_{\text{H}_1(s),\text{H}_2(l)} = \begin{cases} -\text{c}_{12} & \text{if } l \in \text{cl}(s) \\ +\text{c}_{12} & \text{if } l \in \text{cr}(s) \end{cases}, \quad (6)$$

where $\text{c}_{12}$ is a constant. The weights are sign matched according to the routing directions, i.e. negative when $l$ is in the left subtree from $s$ and positive otherwise. Thus, the activation of $\text{H}_2(l)$ is maximized when all split decisions routing to $l$ are satisfied. All other weights are zero.

To encode the leaf to which a data sample $x$ is routed, the bias is set to $b_{\text{H}_2(l)} = -\text{c}_{12} \cdot (|\text{P}(l)| - 1)$ so that the weighted output of neuron $\text{H}_2(l)$ will be greater than zero when all split decisions along the path are satisfied and less than zero otherwise. By using the activation function $\text{a}(\cdot) = sigmoid(\cdot)$, the active neuron $\text{H}_2(l)$ with $x \in l$ will map close to 1 and all other neurons close to 0.

**Output Layer.** The output layer contains one neuron $\text{H}_3(c)$ for each class and is fully-connected to the previous layer $\text{H}_2$. Each neuron $\text{H}_2(l)$ indicates whether $x \in l$. The corresponding leaf node $l$ in the decision tree stores the class votes $y_c^l$ for each class $c$. To transfer the voting system, the weights are set proportional to the class votes:

$$w_{\text{H}_2(l),\text{H}_3(c)} = \text{c}_{23} \cdot y_c^l, \quad (7)$$

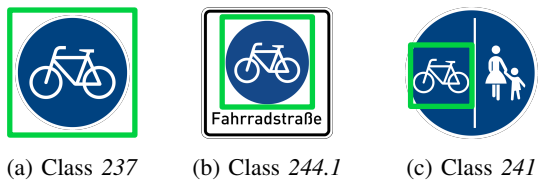

(a) Class *237*  (b) Class *244.1*  (c) Class *241*

Fig. 3: The subject from class *237* (a) occurs similarly in class *244.1* (b) and class *241* (c). Due to very few training examples and the consequent low variability, parts of traffic signs are recognized. We utilize this information and integrate the recognition of parts into the bounding box prediction.

where $c_{23}$ is a scaling constant to normalize the votes as explained in the following section. All biases are set to zero.

**Random Forest.** Extending the mapping to random forests with $T$ decision trees is simply done by mapping each decision tree and concatenating the neurons of the constructed neural networks for each layer. The neurons for each class in the output layer are created only once. They are fully-connected to the previous layer and by setting the constant $c_{23}$ to $1/T$ the outputs of all trees is averaged. We denote the resulting neural network as $\text{NN}^{\text{RF}}$. It should be noted that the memory size of the mapped neural network is linear to the total number of split and leaf nodes. A possible network splitting strategy for very large random forests has been presented by Massiceti *et al.* [19].

### D. Fully Convolutional Network

Mapping the random forest to a neural network allows to join the feature generator and the classifier. Therefore we remove the classification layers from $\text{CNN}^{\text{F}}$, i.e. all layers after *relu3*, and append all layers from $\text{NN}^{\text{RF}}$. The constructed network $\text{CNN}^{\text{F+RF}}$ processes an image patch and outputs the class probabilities.

The convolutional neural network $\text{CNN}^{\text{F+RF}}$ is converted to a fully convolutional network $\text{CNN}^{\text{FCN}}$ by converting the fully-connected layers into convolutional layers, similar as [23]. The fully convolutional network operates on input images of any size and produces corresponding (possibly scaled) output maps. Compared to patch-wise processing, the classifier is naturally slided over the image evaluating the class probabilities at any position. At the same time the features are shared so that features in overlapping patches can be reused. This decreases the amount of computation and significantly accelerates the processing of full images.

### E. Bounding Box Prediction

The constructed fully convolutional network processes a color image $I \in \mathbb{R}^{W \times H \times 3}$ of size $W \times H$ with three color channels and produces an output $O = \text{CNN}^{\text{FCN}}(I)$ with $O \in \mathbb{R}^{W' \times H' \times C}$. The output consists of $C$-dimensional vectors at any position which indicate the probabilities for each class. Due to stride and padding parameters, the size of the output map can be decreased. To detect objects of different sizes, we process the input image in multiple scales $S = \{s_1, \ldots, s_m\}$.

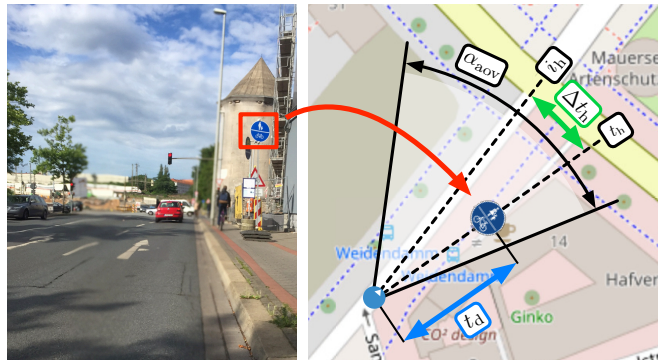

Fig. 4: The detections are projected to the map by integrating additional data. Based on the position $(i_{\text{lat}}, i_{\text{lon}})$ and heading $i_{\text{h}}$ of the image, the position $(t_{\text{lat}}, t_{\text{lon}})$ and heading $t_{\text{h}}$ of the traffic sign are determined. To approximate the geoinformation depending on the position and size of the bounding box, the relative heading $\Delta t_{\text{h}}$ (green) and distance $t_{\text{d}}$ (blue) between the image and traffic sign are calculated.

We extract potential object bounding boxes by identifying all positions in the output maps where the probability is greater than a minimal threshold $t_{\min} = 0.2$. We describe a bounding box by $b = (b_{\text{x}}, b_{\text{y}}, b_{\text{w}}, b_{\text{h}}, b_{\text{c}}, b_{\text{s}})$, where $(b_{\text{x}}, b_{\text{y}})$ is the position of the center, $b_{\text{w}} \times b_{\text{h}}$ the size, $b_{\text{c}}$ the class, and $b_{\text{s}}$ the score. The bounding box size corresponds to the field of view which is equal to the size of a single image patch. All values are scaled according to the scale factor. The score $b_{\text{s}}$ is equal to the probability in the output map.

For determining the final bounding boxes, we process the following three steps. First, we apply non-maximum suppression on the set of bounding boxes for each class to make the system more robust and accelerate the next steps. Therefore we iteratively select the bounding box with the maximum score and remove all overlapping bounding boxes. Second, traffic signs are special classes since the subject of one traffic sign can be included similarly in another traffic sign as illustrated in Fig. 3. We utilize this information by defining a list of parts that can occur in each class. A part is found when a bounding box $b'$ with the corresponding class and an Intersection over Union (IoU) greater than $0.2$ exists. If this is the case we increase the score by $b'_{\text{s}} \cdot 0.2 / P$, where $P$ is the number of parts. Third, we perform non-maximum suppression on the set of all bounding boxes by iteratively selecting the bounding box with the maximum score and removing all bounding boxes with IoU $> 0.5$.

The final predictions are determined by selecting all bounding boxes that have a score $b_{\text{s}}$ greater or equal than a threshold $t_c$ for the corresponding class.

## IV. LOCALIZATION

In this process we integrate additional data from other sensors to determine the position and heading of the traffic signs. For localization of the traffic signs, we use the GPS-position $(i_{\text{lat}}, i_{\text{lon}})$ and heading $i_{\text{h}}$ of the images. The heading is the direction to which a vehicle is pointing. The data is included in our dataset which is described in detail in Section V. As illustrated in Fig. 4, we transform each bounding box $b = (b_x, b_y, b_w, b_h, b_c, b_s)$ to a traffic sign $t = (t_{lat}, t_{lon}, t_h, t_c)$, where $(t_{lat}, t_{lon})$ is the position, $t_h$ the heading, and $t_c$ the class. Since the position and viewing direction of the image is known, we approximate the traffic sign position and heading by calculating the relative heading $\Delta t_h$ and distance $t_d$.

The relative heading is based on the horizontal position $b_x$ of the bounding box in the image. We calculate the horizontal offset to the center of the image normalized by the image width $i_w$. Additionally, we multiply the value by the estimated angle of view $\alpha_{aov}$. Thereby, the relative heading is calculated by $\Delta t_h = \alpha_{aov} \cdot (b_x/i_w - 0.5)$.

The distance $t_d$ between the position of the image and the position of the traffic sign is approximated by estimating the depth of the bounding box in the image. Traffic signs have a defined size $t_w \times t_h$, where $t_w$ is the width and $t_h$ the height. Since an approximate depth estimation is sufficient, we use the information about the size and assume a simple pinhole camera model. Given the focal length $f$ and the sensor width $s_w$ of the camera obtained from the data sheet and a bounding box with width $b_w$, we calculate the approximated distance by $t_d = f \cdot t_w \cdot i_w/(b_w \cdot s_w)$.

Lastly, a traffic sign $t = (t_{lat}, t_{lon}, t_h, t_c)$ is generated. The class $t_c$ equals the bounding box class and the heading is calculated by adding the relative heading to the heading of the image $t_h = i_h + \Delta t_h$. The traffic sign position $(t_{lat}, t_{lon})$ is determined by moving the position of the image by $t_d$ in the direction $t_h$.

## V. DATASET

To collect data in real-world environments, smart phones are used for data recording because they can be readily attached to bicycles. Many people own a smart phone so that a large number of users can be involved. The recorded dataset consists of more than $40\,000$ images.

### A. Data Capturing

We developed an app for data recording which can be installed onto the smart phone. Using a bicycle mount, the smart phone is attached to the bike oriented in the direction of travel. While cycling, the app captures images and data from multiple sensors. Images of size $1080 \times 1920$ pixels are taken with a rate of one image per second. Sensor data is recorded from the built-in accelerometer, gyroscope, and magnetometer with a rate of ten data points per second. Furthermore, geoinformation is added using GPS. The data is recorded as often as the GPS-data is updated.

### B. Filtering

After finishing a tour, the images are filtered to reduce the amount of data. Especially monotonous routes, e.g. in rural areas, produce many similar images. However, the rate with which images are captured cannot be reduced because this increases the risk of missing interesting situations.

We therefore introduce an adaptive filtering of the images. The objective is to keep images of potentially interesting situations that help to analyze traffic situations, but to remove redundant images. For instance, interesting situations could be changes in direction, traffic jams, bad road conditions, or obstructions like construction works or other road users.

For filtering, we integrate motion information and apply a twofold filtering strategy based on decreases in speed and acceleration: (i) **Decreases in speed** indicate situations where the cyclist has to slow down because of potential traffic obstructions such as for example traffic jams, construction works, or other road users. Speed is provided by the GPS-data. We apply a derivative filter to detect decreases in speed. As filter, we use a derivative of Gaussian filter with a bandwidth, i.e. standard deviation, of $2\,\frac{km}{h^2}$. (ii) **Acceleration** is used to analyze the road conditions and to detect for example bumps. It is specified per axis. Each data point consists of a three-dimensional vector. We calculate the Euclidean norm of the vector and apply two smoothing filters with different time spans: One with a large and one with a short time span. Thus, we filter the noisy acceleration data and detect the situations in which the short-term average acceleration relative to the long-term average acceleration exceeds a threshold of $k$. For smoothing, we use Gaussian filters with bandwidths of $1.5\,g$ and $10\,g$, with standard gravitational acceleration $g = 9.81\,\frac{m}{s^2}$, and set $k = 2.8$. We filter the images by removing images if none of the two criteria indicates an interesting situation.

The filtering process reduces the amount of data by a factor of $5$ on average. Subsequently, the data is transfered to a server.

## VI. EXPERIMENTS

Experiments are conducted to demonstrate the performance of the recognition system. Due to the limited amount of labeled data, the pipeline is trained on patches and then extended to perform object recognition. First, results are presented on the classification of a single patch. Afterwards, the recognition performance is illustrated. The comparison of patch-wise processing and fully convolutional processing of full images is shown in the end.

Random forests are trained and tested on a Intel(R) Xeon(R) CPU E5-2650 v3 @2.30GHz, and neural networks on a NVIDIA GeForce GTX Titan X using the *Caffe* [24] framework. The proposed system is programmed in Python.

### A. Training and Test Data

$10$ different traffic signs that are interesting for cyclists are selected. Because the signs differ from traffic signs for cars, the availability of labeled data is very limited. Some classes come with few labeled data but for some classes no labeled data is available.

To have ground truth data of our classes for training and testing, we manually annotated $297$ bounding boxes of traffic signs in the images. The data is split into training set and test set using a split ratio of $50/50$. In Fig. 5, the number of samples per class are shown. The training data consists of $146$ samples for all $10$ classes which corresponds to less single or two misclassified examples. Classes *1000-32* and *242.1* which consist of 4 and 7 examples have larger errors. Classes *242.1* and *244.1* which have a similar appearance are confused once. Some background examples are classified as traffic signs and vice versa. Please confer to Fig. 6 for more information about the traffic signs the classes correspond to.

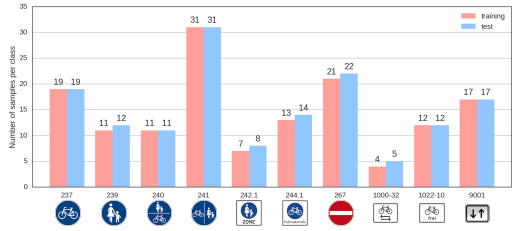

Fig. 5: Number of training and test samples in each class. On average only 15 samples are available per class for each set.

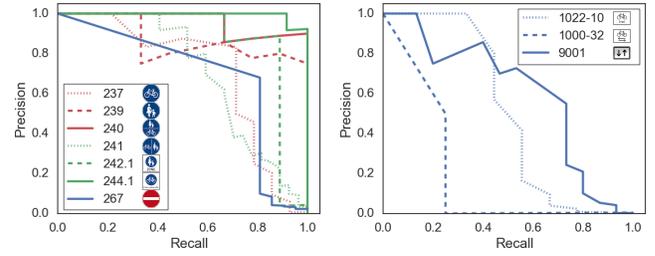

(a) Standard traffic signs  (b) Info signs

Fig. 7: Precision-recall curves for evaluating the recognition performance. The shape of the curves is erratic because few labeled data is available for training and testing.

### C. Object Recognition

The next experiment is conducted to demonstrate the recognition performance of the proposed system. The task is to detect the position, size, and type of all traffic signs in an image. The images have a high diversity with respect to different perspectives, different lighting conditions, and motion blur.

The recognition system is constructed by extending the CNN for patch-wise classification to a fully convolutional network so that fast processing of full images is enabled. A filtering strategy is applied subsequently to predict bounding boxes. No additional training data is required during this process so that only 146 examples over 10 classes are used for training the recognition system. We process the images in 8 different scales. Starting with the scale $s_0 = 1$, the image size is decreased from scale to scale by a factor of $1.3$.

To evaluate the recognition performance, we process all images in the test set and match the predicted bounding boxes with the ground truth data. Each estimated bounding box is assigned to the ground truth bounding box with the highest overlap. The overlap is measured using the IoU and only overlaps with an IoU $> 0.5$ are considered.

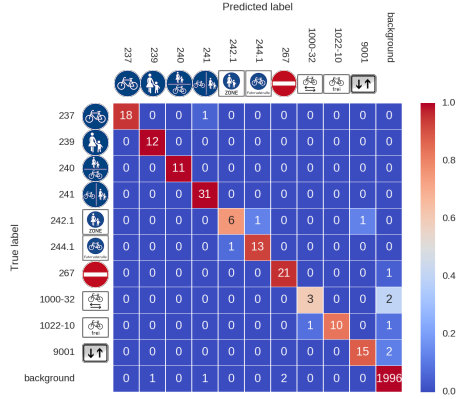

Fig. 6: Confusion matrix showing the performance of the classifier on the test set. The absolute number of samples are shown in the matrix.

than 15 samples per class on average. Please note that class *1000-32* has only 4 examples for training.

Additionally, $2\,000$ background examples are randomly sampled for training and testing. The splitting is repeated multiple times and the results are averaged.

### B. Classification

The first experiment evaluates the performance of the classification on patches. The evaluation is performed in two steps. First, the training for learning features is examined and, secondly, the classification on the target task.

For feature learning, the GTSRB [1] dataset is used since it is similar to our task and has a large amount of labeled data. The dataset consists of $39\,209$ examples for training and $12\,630$ examples for testing over 43 classes. After training, the convolutional neural network $CNN^F$ achieves an accuracy of $97.0\%$ on the test set.

In the next step, the learned features are used to generate a feature vector of each training example of our dataset, and then to train a random forest. For evaluation, the test data is processed similarly. A feature vector is generated for each example from the test set using the learned feature generator $CNN^F$ and classified by the random forest subsequently.

The random forest classification achieves an accuracy of $99.3\%$ on the test set. The confusion matrix is shown in Fig. 6. Three classes are classified without errors. All other classes, except from the background class, only contain a

All bounding boxes come with a score and the class specific threshold $t_c$ determines if a bounding box is accepted or rejected as described in Section III-E. For each class, the threshold $t_c$ is varied, and precision and recall are calculated. The resulting precision-recall curves are shown in Fig. 7. To facilitate understanding these results, two graphs are shown. In the first, the precision-recall curves of a group of standard traffic signs are plotted. The results are good. Some classes are detected almost perfectly. In the second graph, the precision-recall curves of a different group of traffic signs are plotted. These signs are much more difficult to recognize as they are black and white and do not have a conspicuous color. The performance of each class correlates with the number of examples that are available for training. Class *9001* with

| class | 237 | 239 | 240 | 241 | 242.1 | 244.1 | 267 | 1000-32 | 1022-10 | 9001 |
|---|---|---|---|---|---|---|---|---|---|---|
| AP | 0.694 | 0.880 | 0.967 | 0.696 | 0.869 | 0.994 | 0.559 | 0.130 | 0.483 | 0.590 |

TABLE I: Average precision of each class on the test dataset.

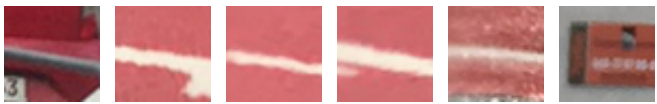

Fig. 8: Selected failure cases for class *267*.

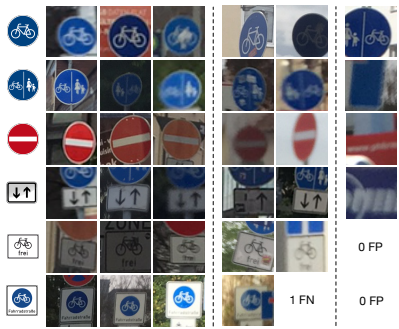

Fig. 9: Recognition results for randomly chosen examples of the test set. In each row, the ground truth traffic sign is shown on the left along with correctly recognized traffic signs (first three columns from the left), false negatives (next two columns), and false positives (last column to the right). Some classes do not have more than a single false negative or no false positives at all.

17 training examples performs best, class *1022-10* with 12 training examples second best, and class *1000-32* with only 4 training examples worst. In Fig. 8 failure cases for class *267* are shown. Patches with similar appearance are extracted due to the limited variability with few training samples and missing semantic information since the broader context is not seen from the patch-wise classifier. To summarize the performance on each class the average precision (AP) is calculated. The results are presented in Table I. In total, the recognition system achieves a good mean average precision (mAP) of 0.686.

In the last step, the final bounding box predictions are determined. The threshold $t_c$ of each class is selected by calculating the F1 score for each precision-recall pair and choosing the threshold with the maximum F1 score. Some qualitative results are presented in Fig. 9. In each row, examples of a particular class are chosen at random. Examples that are recognized correctly are shown in the first three columns, examples that are not recognized are shown in the next two columns. Some of these examples are twisted or covered by stickers. Examples which are recognized as traffic sign but in fact belong to the background or a different class are shown in the column to the right. These bounding box patches can have a similar color or structure.

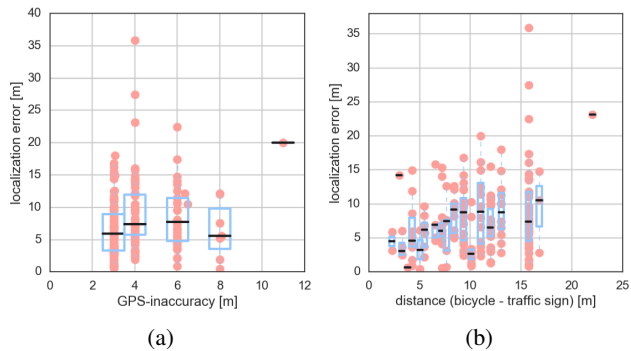

Fig. 10: The distance error with respect to GPS-inaccuracy (a) and distance between the recording device and the traffic sign (b). The black lines indicate the medians, the upper and bottom ends of the blue boxes the first and third quantile.

### D. Computation Time

In the third experiment we evaluate the computation time. Random forests are fast at test time for the classification of a single feature vector. When processing a full image, the random forest is applied to every patch in the feature maps. For an image of size $1080 \times 1920$ the feature maps are produced relatively fast using $\text{CNN}^\text{F}$ and have a size of $268 \times 478$ so that 124 399 patches have to be classified to build the output probability map. The images are process in 8 different scales. All together, we measured an average processing time of more than 10 hours for a single image. Although, the computation time could be reduced by using a more efficient language than Python, the time to access the memory represents a bottleneck due to a large overhead for accessing and preprocessing each patch.

For processing all in one pipeline, we constructed the fully convolutional network $\text{CNN}^\text{FCN}$. The network combines feature generation and classification and processes full images in one pass. The time for processing one image in 8 different scales is only 6.08 seconds on average. Compared to the patch-wise processing using random forest, using the fully convolutional network reduces the processing time significantly.

### E. Precision of Localizations

The last experiment is designed to demonstrate the localization performance. The localization maps the predicted bounding boxes in the image to positions on the map. Position and heading of a traffic sign are calculated based on the geoinformation of the image as well as the position and size of the bounding boxes.

For evaluation, we generate ground truth data by manually labeling all traffic signs on the map that are used in our dataset. In the next step, correctly detected traffic signs are matched with the ground truth data. The distance between two GPS-positions is calculated using the *haversine formula* [25]. The maximal possible difference of the heading is $90°$ because larger differences would show a traffic sign from the side or from the back. Each traffic sign is assigned to the ground truth traffic sign that has the minimum distance and a heading difference within the possible viewing area of $90°$. The median of the localization error, i.e. the distance between the estimated position of the traffic sign and its ground truth position, is $6.76 \, \text{m}$. Since the recorded GPS-data also includes the inaccuracies of each GPS-position, we can remove traffic signs which are estimated by more inaccurate GPS-positions. If traffic signs with a GPS-inaccuracy larger than the average of $3.95 \, \text{m}$ are removed, then the median of the localization error decreases to $5.95 \, \text{m}$.

The errors of the localizations (y-axis) with respect to the GPS-inaccuracies (x-axis) are plotted in Fig. 10a. The orange dots indicate estimated positions of traffic signs. The black lines indicate the medians, the upper and bottom ends of the blue boxes the first and third quantiles. It can be seen that the localization error does not depend on the precision of the GPS-position as it does not increase with the latter. The localization errors (y-axis) with respect to the distances between the positions of the traffic signs and the GPS-positions (x-axis) are shown in Fig. 10b. It can be seen that the errors depend on the distance between traffic sign and bicycle as they increase with these distances. This can be explained by the fact that the original inaccuracies of the GPS-position are extrapolated, i.e. the larger the distances, the more the GPS-inaccuracies perturb the localizations.

Since smart phones are used as recording device, the precision of the GPS-coordinates is lower than those used in GPS-sensors integrated in cars or in high-end devices. As the inaccuracies of the GPS-positions have a large influence on the localizations, we intend to identify multiple observations of the same sign in future work. Then, the localization error could be reduced by considering multiple observations of the same traffic sign.

## VII. CONCLUSION

We presented a system for object recognition that is trained with very few labeled data. CNNs have shown great results in feature learning and random forests are able to build a robust classifier even if little data is available. We combined the advantages of CNNs and random forests to construct a fully convolutional network for predicting bounding boxes.

The system is built in three steps. First, we learned features using a CNN and trained a random forest to perform patch-wise classification. Second, the random forest is mapped to a neural network. Afterwards, we transform the pipeline to a fully convolutional network to accelerate the processing of full images. Whereas deep learning typically depends on the availability of large datasets, the proposed system significantly reduces the required amount of labeled data.

The proposed system was evaluated on crowdsourced data with the aim of collecting traffic information for cyclists. We used our system to recognize traffic signs that are relevant for cyclists. The system is trained with only 15 examples per class on average. Furthermore, we showed how additional sensor information can be used to locate traffic signs on the map.

## REFERENCES

[1] J. Stallkamp, M. Schlipsing, J. Salmen, and C. Igel, "Man vs. computer: Benchmarking machine learning algorithms for traffic sign recognition," *Neural Networks*, 2012.
[2] A. Geiger, P. Lenz, C. Stiller, and R. Urtasun, "Vision meets robotics: The kitti dataset," *IJRR*, 2013.
[3] M. Cordts, M. Omran, S. Ramos, T. Rehfeld, M. Enzweiler, R. Benenson, U. Franke, S. Roth, and B. Schiele, "The cityscapes dataset for semantic urban scene understanding," in *CVPR*, 2016.
[4] A. Krizhevsky, I. Sutskever, and G. E. Hinton, "Imagenet classification with deep convolutional neural networks," in *NIPS*, 2012.
[5] F. Kluger, H. Ackermann, M. Y. Yang, and B. Rosenhahn, "Deep learning for vanishing point detection using an inverse gnomonic projection," in *GCPR*, Sept. 2017.
[6] M. Y. Yang, W. Liao, H. Ackermann, and B. Rosenhahn, "On support relations and semantic scene graphs," *ISPRS Journal of Photogrammetry and Remote Sensing*, vol. 131, pp. 15–25, July 2017.
[7] R. Girshick, J. Donahue, T. Darrell, and J. Malik, "Rich feature hierarchies for accurate object detection and semantic segmentation," in *Computer Vision and Pattern Recognition*, 2014.
[8] K. He, X. Zhang, S. Ren, and J. Sun, "Spatial pyramid pooling in deep convolutional networks for visual recognition," in *European Conference on Computer Vision*, 2014.
[9] R. Girshick, "Fast R-CNN," in *ICCV*, 2015.
[10] S. Ren, K. He, R. Girshick, and J. Sun, "Faster R-CNN: Towards real-time object detection with region proposal networks," in *NIPS*, 2015.
[11] W. Liu, D. Anguelov, D. Erhan, C. Szegedy, S. Reed, C.-Y. Fu, and A. C. Berg, "SSD: Single shot multibox detector," in *ECCV*, 2016.
[12] J. Redmon and A. Farhadi, "Yolo9000: Better, faster, stronger," *arXiv preprint arXiv:1612.08242*, 2016.
[13] M. Oquab, L. Bottou, I. Laptev, and J. Sivic, "Learning and transferring mid-level image representations using convolutional neural networks," in *CVPR*, 2014, pp. 1717–1724.
[14] B. Moysset, C. Kermorvant, and C. Wolf, "Learning to detect and localize many objects from few examples," *CoRR*, vol. abs/1611.05664, 2016.
[15] R. Wagner, M. Thom, R. Schweiger, G. Palm, and A. Rothermel, "Learning convolutional neural networks from few samples," in *IJCNN*, Aug 2013, pp. 1–7.
[16] P. Kontschieder, M. Fiterau, A. Criminisi, and S. R. Bul, "Deep neural decision forests," in *ICCV*, 2015, pp. 1467–1475.
[17] I. K. Sethi, "Entropy nets: from decision trees to neural networks," *Proceedings of the IEEE*, vol. 78, no. 10, pp. 1605–1613, Oct 1990.
[18] J. Welbl, "Casting random forests as artificial neural networks (and profiting from it)," in *GCPR*. Springer, Cham, 2014, pp. 765–771.
[19] D. Massiceti, A. Krull, E. Brachmann, C. Rother, and P. H. S. Torr, "Random forests versus neural networks - what's best for camera localization?" in *ICRA*. IEEE, 2017, pp. 5118–5125.
[20] D. L. Richmond, D. Kainmueller, M. Y. Yang, E. W. Myers, and C. Rother, "Relating cascaded random forests to deep convolutional neural networks for semantic segmentation," *CoRR*, vol. abs/1507.07583, 2015.
[21] J. T. Springenberg, A. Dosovitskiy, T. Brox, and M. Riedmiller, "Striving for Simplicity: The All Convolutional Net," *ICLR*, pp. 1–14, 2015.
[22] L. Breiman, "Random forests," *Machine Learning*, vol. 45, no. 1, pp. 5–32, Oct 2001.
[23] E. Shelhamer, J. Long, and T. Darrell, "Fully convolutional networks for semantic segmentation," *IEEE Trans. Pattern Anal. Mach. Intell.*, vol. 39, no. 4, pp. 640–651, 2017.
[24] Y. Jia, E. Shelhamer, J. Donahue, S. Karayev, J. Long, R. Girshick, S. Guadarrama, and T. Darrell, "Caffe: Convolutional architecture for fast feature embedding," *arXiv preprint arXiv:1408.5093*, 2014.
[25] J. Inman, *Navigation and Nautical Astronomy, for the Use of British Seamen*. F. & J. Rivington, 1849.